\pdfoutput=1
\documentclass[11pt]{article}
\usepackage{eamt23}
\usepackage{times}
\usepackage{url}
\usepackage{latexsym}
\usepackage[small,bf]{caption} 
\usepackage{algorithm}
\usepackage{algorithmicx}
\usepackage{algpseudocode}

\usepackage{booktabs}
\usepackage{tabularx}
\usepackage{array}
\usepackage{graphicx}
\usepackage{float}
\usepackage{subfigure}
\usepackage{amssymb}
\usepackage[mathscr]{eucal}
\usepackage[T1]{fontenc}
\usepackage[utf8]{inputenc}
\usepackage{microtype}
\usepackage{multicol}
\usepackage{multirow}

\title{How to Design Translation Prompts for ChatGPT: An Empirical Study}

\author{Yuan Gao , Ruili Wang , Feng Hou \\
  School of Mathematical and Computational Science \\
  Massey University, New Zealand \\
  \texttt{\{y.gao, ruili.wang, f.hou\}@massey.ac.nz} \\}
  
\date{}

\begin{document}
\maketitle

\begin{abstract}
The recently released ChatGPT has demonstrated surprising abilities in natural language understanding and generation. Given that machine translation heavily relies on these abilities, there is substantial promise in applying ChatGPT for machine translation. Using naive prompts cannot fully unleash ChatGPT's translation ability. Thus, in this paper, we propose several translation prompts that contain (i). translation task information (e.g., English-to-German), (ii). context domain information (e.g., News), (iii). Part-of-Speech (POS) tags, respectively, aiming at further unleashing the translation power of ChatGPT. Our experimental results show that our proposed translation prompts can significantly enhance ChatGPT's performance in translation. We evaluate the translation quality using multi-reference test sets which consist of ten different human translations for each source sentence, and ChatGPT achieves superior performance compared to commercial systems. In addition, we also develop few-shot prompts upon our proposed translation prompts, which consistently show improvement across different translation directions.
\end{abstract}

\section{Introduction}

Machine translation (MT) aims to automatically translate text from one language to another target language with the aid of computers. As an important research area in Artificial Intelligence, MT has attracted massive attention in both academia and industry \cite{stahlberg2020neural,yang2020survey}. In recent years, there has been a growing trend towards incorporating large-scale pre-trained language models for natural language processing \cite{yang2019xlnet,brown2020language}. These models are typically trained on massive text data and are enabled to capture rich representations of the input text. Incorporating large language models has demonstrated substantial advancements in various natural language processing tasks, including machine translation \cite{lewis2020bart}.

The recently released ChatGPT\footnote{\url{https://openai.com/blog/chatgpt/}} is a powerful pre-trained language model developed by OpenAI. ChatGPT is built upon GPT-3.5 and optimized with Reinforcement Learning from Human Feedback (RLHF). Due to its surprising ability of natural language understanding and generation, ChatGPT has attracted millions of users. While ChatGPT is primarily designed as an intelligent conversational system, it can also perform plenty of human-like tasks (e.g., writing poems, fixing coding bugs and so on). However, recent work \cite{jiao2023chatgpt} reveals that ChatGPT with naive prompts exhibits a non-negligible performance gap in comparison with other commercial translation systems, such as Google Translate and DeepL Translate.

Different from other commercial translation systems, ChatGPT is capable of adjusting its output bias based on the provided prompt. As a result, instead of solely requesting translations from ChatGPT, users are allowed to input various translation prompts into its dialogue box with the source input. Since OpenAI only provides a web interface to access ChatGPT, we cannot modify the internal components or retrieve the intermediate representations of ChatGPT. Thus, we treat ChatGPT as a black-box system and explore what kind of translation prompts could fully unleash the translation power of ChatGPT.  We hypothesize that providing translation task or context domain information could make ChatGPT focus on the current input data, thereby improving the quality of translations. Moreover, we also try to introduce Part-of-Speech(POS) tags into prompts as auxiliary information to assist ChatGPT. 

In this work,  we propose several translation prompts that contain (i). translation task information (e.g., English-to-German), (ii). context domain information (e.g., News), (iii). Part-of-Speech (POS) tags, respectively, aiming at further unleashing the translation power of ChatGPT. We conduct comprehensive experiments to evaluate the efficacy of our proposed translation prompts for enhancing the translation performance of ChatGPT.  The experimental results indicate that ChatGPT with our translation prompts significantly outperforms our baseline prompt on multilingual language translations. We evaluate ChatGPT on the multi-reference and multi-domain scenario. We observe significant improvement in performance when measured using multi-reference test sets which consist of ten different human translations for each source sentence, surpassing even commercial translation systems. In the case of multi-domain translations, we analyze the performance through experiments conducted on four different domain test sets. We observe that ChatGPT performs well when using the translation prompt that contains domain information and consistently achieves higher BLEU scores than other prompts. Additionally, we provide wrong domain information to ChatGPT to verify if it can correctly understand domain information. The experimental results demonstrate that ChatGPT is able to comprehend the provided domain information and subsequently adjust the generated translation accordingly.

At last, we develop few-shot prompts upon our proposed translation prompts. This approach is motivated by previous research that utilize large-scale language models (LLMs) to enhance downstream tasks by providing several input-output examples \cite{brown2020language,chen2021revisiting,dou2022gpt}. To mitigate any unexpected effects on ChatGPT, we selectively sample multiple high-quality translation pairs from the same dataset as examples. Note that the selected examples and test samples are mutually exclusive, ensuring reliable and unbiased evaluation.

\section{Background}

\textbf{Machine Translation(MT)} is a crucial area of research in the field of natural language processing and has gained immense attention in recent years. The primary goal of MT is to automatically translate text data from the source language to the target language with the help of computers. Thus, a mature translation system must possess robust language understanding and language generation capabilities in order to produce adequate and fluent translations. Previous works \cite{liu2019incorporating,guo2020incorporating} have shown that LLMs can enhance the ability of translation systems to understand the source text but struggle to improve generation ability. ChatGPT has exhibited an impressive level of performance in both natural language understanding and natural language generation, as evidenced by its ability to comprehend and generate human-like responses in a wide range of contexts. Consequently, exploring the application of ChatGPT in translations presents a compelling and promising area of research. \\

\noindent \textbf{ChatGPT} is a large-scale language model that was fine-tuned upon GPT-3.5. As the official website states, ChatGPT was optimized using RLHF and trained on massive amounts of text data to generate a detailed response by following instructions provided in a prompt. While ChatGPT is primarily designed as an intelligent conversational system, it is capable to perform various human-like tasks, including machine translation. However, recent work \cite{jiao2023chatgpt} found that ChatGPT exhibits an undeniable performance gap when compared with other commercial translation systems, such as Google Translate and DeepL Translate, and this gap is exacerbated in low-resource languages. Thus, we will explore how to unleash the power of ChatGPT for translations by designing different translation prompts.

Inspired by the training objective of ChatGPT, which employs prompts to guide the generation of responses, we consider that a proper translation prompt is able to unleash more translation potential of ChatGPT. In this work, we adopt several translation prompts to trigger ChatGPT for translation and evaluate them in comprehensive experiments.

\section{Experiments}
\begin{table}[t]
\small

\begin{tabular}{lp{6cm}}
\toprule[2pt]
\multicolumn{2}{c}{\textbf{Translation Prompt}}                                                                                                                          \\ 
\midrule[1pt]
TP3                        & {\ttfamily Please provide the [TGT] translation for these sentences:}                                                                                   \\
TT                         & {\ttfamily This is a [SRC] to [TGT] translation, please provide the [TGT] translation for these sentences:}                                        \\
CD                         & {\ttfamily Please provide the [TGT] translation for these sentences taken from the [SPECIFIC DOMAIN]:}                                                  \\ 
\hline
TT-pos                     & {\ttfamily This is a [SRC] to [TGT] translation, please provide the [TGT] translation for these sentences with the help of given POS tags:}        \\
CD-pos                     & {\ttfamily Please provide the [TGT] translation for these sentences taken from the [SPECIFIC DOMAIN] with the help of given POS tags:}                  \\ 
\bottomrule[2pt]
\end{tabular}
\caption{Translation prompts adopted in this work. TT stands for translation task, CD stands for Context Domain, "-pos" indicates that the prompt contains a POS tag. TP3 is extracted from Jiao \shortcite{jiao2023chatgpt} directly without any editions, and other prompts are built upon TP3.}
\label{tab1}
\end{table}

In this section, we first describe our designed translation prompts and provide details on the experimental setup, including the datasets, baselines and evaluation metrics used. We also conduct various experiments to explore the effectiveness of the proposed prompts under different translation tasks. The experimental results and analysis are presented as well.

\subsection{Prompt Design}

\begin{table*}[h]
\centering
\normalsize
\begin{tabular}{l|ccc|ccc|ccc}
\hline
\multirow{2}{*}{\textbf{System}} & \multicolumn{3}{c|}{\textbf{En-\textgreater{}Fr}}      & \multicolumn{3}{c|}{\textbf{Fr-\textgreater{}En}}        & \multicolumn{3}{c}{\textbf{Fr-\textgreater{}Es}}          \\ \cline{2-10}
         & {\small BLEU $\uparrow$}  & {\small ChrF++$\uparrow$} & {\small TER$\downarrow$}             & {\small BLEU $\uparrow$}  & {\small ChrF++$\uparrow$} & {\small TER$\downarrow$}             & {\small BLEU $\uparrow$}  & {\small ChrF++$\uparrow$} & {\small TER$\downarrow$}   \\ \hline
Google                           &  \textbf{54.75}  &  \textbf{75.10}  &  \textbf{33.64}  &  49.66            &  72.74            &  35.48            &  22.48            &  48.50            &  64.76            \\
DeepL                            &  53.87           &  74.65           &  \textbf{33.64}  &  50.09            &  72.29            &  35.30            &  21.68            &  47.67            &  65.38            \\ \hline
TP3                              &  45.03           &  69.50           &  42.11           &  44.86            &  68.85            &  36.02            &  21.33            &  48.25            &  65.31            \\
TT                               &  45.85           &  70.22           &  40.89           & \textbf{50.18*}   & \textbf{72.84*}   &  36.76            &  21.92            &  48.53            &  65.24            \\
CD                               &  45.92           &  70.40           &  40.67           &  49.68            &  72.72            & \textbf{35.05*}   &  22.13            &  48.97            &  64.28            \\
\hline
TT - pos                         &  47.11*          &  71.05*          &  39.40*          &  48.87            &  72.39            &  36.08            &  22.20            &  48.88            &  65.10            \\
CD - pos                         &  46.93           &  70.31           &  39.60           &  49.33            &  72.42            &  36.84            &  \textbf{22.58*}  &  \textbf{49.25*}  &  \textbf{63.79*}  \\
\hline \hline
\multirow{2}{*}{\textbf{System}} & \multicolumn{3}{c|}{\textbf{En-\textgreater{}Es}}      & \multicolumn{3}{c|}{\textbf{Es-\textgreater{}En}}        & \multicolumn{3}{c}{\textbf{Es-\textgreater{}Fr}}          \\ \cline{2-10}
         & {\small BLEU $\uparrow$}  & {\small ChrF++$\uparrow$} & {\small TER$\downarrow$}             & {\small BLEU $\uparrow$}  & {\small ChrF++$\uparrow$} & {\small TER$\downarrow$}             & {\small BLEU $\uparrow$}  & {\small ChrF++$\uparrow$} & {\small TER$\downarrow$}   \\ \hline
Google                           &  23.49           &  50.73           &  62.62           &  25.32           &  53.21           &  70.01              &  26.89           &  53.47           &  66.07              \\
DeepL                            &  23.62           &  49.86           &  63.59           &  26.74           &  54.64           &  67.44              &  \textbf{29.55}  &  \textbf{54.19}  &  \textbf{61.84}     \\ \hline
TP3                              &  23.00           &  49.93           &  62.28           &  25.42           &  54.07           &  70.35              &  26.20           &  53.23           &  65.85              \\
TT                               &  23.47           &  50.43           &  62.69           & \textbf{27.10*}  & \textbf{55.40*}  &  67.81              &  26.66*          &  53.92*          &  63.72*             \\
CD                               &  23.33           &  50.76           &  62.69           &  26.89           &  54.91           & \textbf{67.10*}     &  26.05           &  53.58           &  64.00              \\
\hline
TT - pos                         &  23.69           &  50.81           &  63.17           &  26.42           &  53.84           &  67.52              &  26.35           &  53.63           &  64.15              \\
CD - pos                         & \textbf{23.92*}  & \textbf{51.06*}  & \textbf{62.07*}  &  26.37           &  54.64           &  68.21              &  26.19           &  53.38           &  63.93              \\
\hline
\end{tabular}
\caption{Performance comparison of ChatGPT for high-resource translations using BLEU, ChrF++ and TER scores. The best scores are highlighted in bold, and * denotes the highest score achieved by ChatGPT across different prompts.}
\label{tab2}
\end{table*}

ChatGPT is trained to generate corresponding responses according to the provided prompts. This nature determines that it is highly sensitive to the information provided by prompts. Jiao \shortcite{jiao2023chatgpt} used three different prompts, provided by ChatGPT itself, which show similar translation performance. The prompts are as follows:

\begin{quote}
\small
\begin{itemize}
\item {\ttfamily TP1 - Translate these sentences from [SRC] to [TGT]:}
\item {\ttfamily TP2 - Answer with no quotes. What do these sentences mean in [TGT]?}
\item {\ttfamily TP3 - Please provide the [TGT] translation for these sentences:}
\end{itemize}
\end{quote}

We argue that the translation results are unlikely to be significantly impacted by the three prompts, as they merely provide general instructions advised by ChatGPT without any supplementary information. It is worth noting that ChatGPT was trained on massive amounts of multilingual general-domain datasets, and as a result, the generated text is inevitably influenced by various prior knowledge instead of being solely based on the current input. Thus, we consider specifying translation task or context domain could make ChatGPT more focused on the current input text, resulting in better translations. In light of this, we propose two translation prompts, named TT and CD, from two perspectives and compare them with the most effective prompts of \cite{jiao2023chatgpt} (refer to TP3). Besides, we integrated POS tags as auxiliary information into both TT and CD, named TT-pos and CD-pos, respectively. All translation prompts adopted in this work are presented in Table \ref{tab1}.

\begin{figure}[t]
\includegraphics[width=0.5\textwidth]{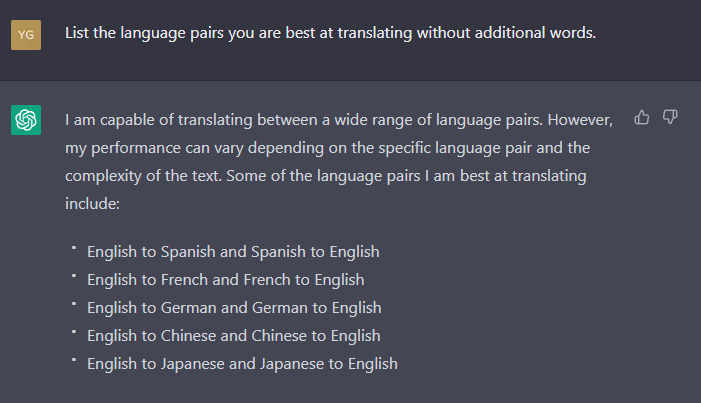}
\caption{Best performing translations answered by ChatGPT.}
\label{fig1}
\end{figure}

\subsection{Experimental Setup}

\subsubsection{Datasets}
We conduct experiments on a range of benchmarks, including multilingual translations, multi-reference translations and multi-domain translations. For the multilingual scenario, we choose English $\leftrightarrow$ Spanish and English $\leftrightarrow$ French, which ChatGPT can perform high-quality translation as shown in Figure \ref{fig1}, and we further conduct Spanish $\leftrightarrow$ French translations to evaluate ChatGPT in the non-English-centric scenario. We use the Flores-101 dataset \cite{goyal2022flores} for all of the above translation directions, which contains 1012 sentences from Wikipedia articles for each language. Since the text generated by ChatGPT is more flexible and diverse than that of conventional translation systems, it can be challenging to evaluate ChatGPT's actual performance with a single reference. Consequently, we employ multi-reference test sets which collect 10 different human translations for each source sentence to evaluate the translation quality of ChatGPT, and the data is released by \cite{ott2018analyzing}. 

Due to the response delay caused by the available computing resources, using ChatGPT for translation can be a time-consuming and labour-intensive process. Therefore, we follow the strategy outlined in \cite{jiao2023chatgpt} to randomly sample 50 sentence pairs from each test set for our final test sets.

\subsubsection{Baselines and Evaluation Metrics}

To evaluate the efficacy of our proposed prompts, we compare them against TP3, which serves as a baseline in which no supplementary information is provided about the input text. In addition, as ChatGPT is a well-established pre-trained system, we complement our study by comparing its performance over different prompts with two dominant commercial translation systems, Google Translate and DeepL Translate, instead of the academic translation systems that are trained from scratch. We access ChatGPT through the publicly released web interface\footnote{\url{https://chat.openai.com/chat}}, and we access Google Translate and DeepL Translate in the same way. To ensure the reliability of results, we take the average score obtained from three randomly sampled test sets to ensure that our evaluation results are not impacted by randomness. Furthermore, we evaluate the performance of our system using multiple metrics, including BLEU \cite{papineni2002bleu}, ChrF++ \cite{popovic2017chrf++} and TER \cite{snover2006study}. In this work, we report BLEU scores using SacreBLEU \cite{post2018call}.

\subsection{Multilingual Translation} 

We conduct a complete multilingual translation among three high-resource languages, English(En), French(Fr) and Spanish(Es), all of which use Latin script and belong to the European language family (English belongs to Germanic, French and Spanish belong to Romance \cite{fan2021beyond}). The results are reported in Table \ref{tab2}. It can be observed that the performance of ChatGPT with TP3 generally falls behind that of either Google Translate or DeepL Translate, but remains competitive in certain translation directions, which is consistent with the experimental results of \cite{jiao2023chatgpt,peng2023towards}. Specifically, ChatGPT obtains a +0.10 higher BLEU score than Google Translate on the Spanish $\rightarrow$ English translation. 

Compared to the results of TP3, our translation prompts improve the BLEU scores by 0.89, 5.32, 0.47 and 1.68 BLEU points on four English-centric translations respectively, which demonstrate the efficacy of our prompts. The surprising results appear on the X $\rightarrow$ English translations, where our prompts significantly outperformed TP3 and even exceeded the commercial systems by 0.52 and 1.78 BLEU scores in French $\rightarrow$ English and Spanish $\rightarrow$ English translations, respectively. The findings suggest that ChatGPT has significant potential for English generation tasks. As for Spanish $\leftrightarrow$ French translations, our prompts yield only marginal improvement over TP3, unlike the surprising gains on the English-centric tasks. 




\subsection{POS tags for ChatGPT}

Neural Machine Translation has been shown to benefit from additional provided linguistic features, especially for underrepresented languages. Petrushkov et al. \shortcite{petrushkov2018learning} use human feedbacks to improve the accuracy and robustness of translation. Niu and Carpuat \shortcite{niu2020controlling} adopt target language formality in different ways to improve the performance of the translation model. Li et al. \shortcite{li2022prompt} insert word constraints and syntax constraints into prompts to improve translation quality. In addition, extensive works \cite{khan2020novel,perera2022improving} were conducted on the utilization of POS tags for improving translations. POS tags provide syntactic information that locates each word in a sentence to help disambiguate words and understand the grammatical structure of a sentence. For example, knowing the POS tag of a word can help determine whether it is a subject, object, or modifier in a sentence. Therefore, we conjecture that incorporating POS tags as auxiliary information could further help ChatGPT release translation ability upon TT and CD. We concatenate POS tag sequences with the original sentences directly and prepend $[sentence]/[POS]$ tags in two sequences to identify and segment them. We apply Stanza \cite{qi2020stanza} to automatically generate POS tags for test samples. 

As shown by the results in Table \ref{tab2}, we can see that incorporating POS tags boost ChatGPT in many translation directions, such as improving 
English $\rightarrow$ French with +2.08 BLEU points, French $\rightarrow$ Spanish with +1.25 BLEU points and English $\rightarrow$ Spanish with +0.92 BLEU points. However, the performance drops in some directions, including French $\rightarrow$ English, Spanish $\rightarrow$ English and Spanish $\rightarrow$ French. We suspect that the performance drop is caused by the auxiliary information negatively impacting the original input, either by introducing noise or by increasing the complexity of the input text. While the necessity of incorporating POS tags is not explicitly demonstrated, it is no doubt that auxiliary information has a positive effect on ChatGPT somehow. Furthermore, what kind of auxiliary information and how to integrate are still important research questions for future research. 

Although ChatGPT with POS tags failed to achieve consistent improvement, it shows the ability to fully understand the input content to distinguish between two sequences, and the final output only contains the translation of $[sentence]$ part. In this respect, ChatGPT demonstrates remarkable proficiency.

\subsection{Translation Diversity}

\begin{table}[]
\centering
\begin{tabular}{lcccc}
\toprule[2pt]
\multirow{2}{*}{\textbf{System}} & \multicolumn{2}{c}{\textbf{Single}} & \multicolumn{2}{c}{\textbf{Multiple}} \\ \cline{2-5} 
                                 & BLEU     & $\triangle$              & BLEU      & $\triangle$     \\ \hline
                                 & \multicolumn{4}{c}{\small{En$\rightarrow$Fr}}                                 \\
DeepL                            & 47.46    & -                        & 82.28     & -               \\
Google                           & 49.39    & + 1.93                    & 88.22     & + 5.94         \\
TP3                              & 45.63    & - 1.83                    & 85.59     & + 3.31         \\
TT                               & 46.48    & - 0.98                    & 87.45     & + 5.17         \\
CD                               & 48.65    & + 1.19                    & 87.95     & + 5.67         \\ \hline
                                 & \multicolumn{4}{c}{\small{En$\rightarrow$De}}                                 \\
DeepL                            & 34.88    & -                        & 80.90     & -               \\
Google                           & 38.18    & + 3.30                    & 80.06     & - 0.84         \\
TP3                              & 32.21    & - 2.67                    & 80.56     & - 0.34         \\
TT                               & 33.21    & - 1.67                    & 82.22     & + 1.32         \\
CD                               & 34.28    & - 0.60                    & 82.88     & + 1.98         \\
\bottomrule[2pt]
\end{tabular}
\caption{BLEU scores on En$\rightarrow$Fr and En$\rightarrow$De test sets with single and multiple references.}
\label{tab4}
\end{table}

Since ChatGPT is trained on huge amounts of text data, the model distribution is too widespread in hypothesis space. Thus, the model shows greater flexibility in vocabulary choice during generation. In addition, ChatGPT has been specifically tailored as an intelligent chatbot and has been optimized via the RLHF training scheme, which allows it to produce coherent and concise responses to facilitate reader understanding. In the absence of intervention, however, these attributes of ChatGPT lead to uncertainty translations, which are intentionally generated to enhance fluency and comprehensibility, but inconsistencies with the gold references. 

Therefore, we follow the work of Ott et al. \shortcite{ott2018analyzing} to evaluate the translation quality using multi-reference test sets which collect 10 different human translations for each source sentence. The source sentences of two multi-reference test sets are randomly selected from the WMT 2014 English-French and English-German News sets, respectively. Experimental results are listed in Table \ref{tab4}. Similar to the previous experimental results, ChatGPT still shows a competitive level of performance when evaluated with single references(outperforming DeepL by 1.19 BLEU with CD on En-Fr), while achieving significant improvement when measured by multiple references(surpassing DeepL by 5.67 BLEU with CD on En-Fr). The empirical evidence provided by these results suggests that the utilization of more comprehensive criteria is imperative for the evaluation of ChatGPT's translation performance. Additionally, ChatGPT with CD consistently outperforms other prompts on News sets, contrary to previous results on Flores-101. To gain further insights into this, we conduct multi-domain translation in the next section.

\subsection{Multi-domain Translation} \label{sec3.7}

\begin{table}[t]
\centering
\begin{tabular}{lccc}
\toprule[2pt]
{\small Domains}     & {\small Uni-Words} (\%) & {\small \# of Tokens} & {\small \# of Chars} \\ \hline
{\small News}        & 20.4                    & 20.8                  & 109.4       \\
{\small e-Com.}      & 20.8                    & 22.4                  & 119.4       \\
{\small Social}      & 10.8                    & 19.5                  & 93.2        \\
{\small Conver.}     & 6.7                     & 13.4                  & 59.9         \\ 
\bottomrule[2pt]
\end{tabular}
\caption{The statistical results of adopted domain-specific sets.}
\label{tab5}
\end{table}

The translation performance of ChatGPT exhibits inconsistency when using TT and CD across FLORES (as shown in Table \ref{tab2}) and WMT (as shown in Table \ref{tab4}) test sets. Specifically, ChatGPT with TT performs better than with CD on the FLORSE test sets, whereas using CD is better than using TT on the WMT test sets. This variability in performance may be attributed to the different sources of evaluation data. The FLORES dataset comprises Wikipedia articles that cover multiple topics \cite{goyal2022flores}, and should be considered as a general domain dataset. On the other hand, the data used in Table \ref{tab4} is from a specific (News) domain. To investigate whether ChatGPT with CD is effective for domain-specific translation, we conduct experiments on the multi-domain test sets taken from \cite{hendy2023good}, which cover four domains, namely News, e-Commerce, Social and Conversational. 

Table \ref{tab5} illustrates the diversity in statistical patterns observed in datasets obtained from different domains. The data from different domains display noticeable discrepancies in the proportion of unique words, average sentence length, as well as average word length. In particular, the Conversational domain appears to be a general domain, which contains only 6.7\% of unique words that are not present in other sets. Additionally, the average sentence length and the average number of characters per word are notably shorter in comparison to other domain sets, as shown in Table \ref{tab5}. It is important to note that we employ the lowercase and lemmatization of words before counting them, which is aim to eliminate the impact of morphological transformation on vocabulary overlap, and the calculation of the proportion is based on the total number of tokens rather than the number of words in the vocabulary. 

\begin{table}[t]
\centering
\small
\begin{tabular}{lccccc}
\toprule[2pt]
Prompts & News  & e-Com. & Social & Conver. & Avg. \\ \midrule[1pt]
TP3     & 35.60          & 25.47           & 34.59           & 34.93            & 32.65           \\\midrule[0pt]
TT      & 37.73          & 26.87           & 35.10           & 35.19            & 33.72           \\\midrule[0pt]
CD      & \textbf{39.93} & \textbf{30.28}  & \textbf{35.80}  & \textbf{36.25}   & \textbf{35.57}         \\\midrule[1pt]
w-CD    & 38.03          & 25.41           & 35.13           & 36.04            & 33.80           \\ 
\bottomrule[2pt]
\end{tabular}
\caption{BLEU scores on multi-domain translation. The best scores are highlighted in bold.}
\label{tab6}
\end{table}

As shown in Table \ref{tab6}, ChatGPT with CD achieves remarkable performance across all domains and gains an average BLEU score higher than TP3 and TT by +2.92 and +1.85, respectively. Meanwhile, the News and e-Commerce domains with higher unique word proportions achieve greater improvements (outperforming TP3 by +4.33 and +4.81, respectively) compared to the Social and Conversational domains (outperforming TP3 by +1.21 and +1.32, respectively). This suggests that domain information is advantageous for ChatGPT in translating data from specific domains. However, the results obtained from the BLEU scores demonstrated a lack of apparent correlation between translation quality and domain uniqueness. To assess the impact of providing incorrect domain information, we conducted a comparative experiment using the wrong domain information to prompt ChatGPT (referred to as w-CD). The experimental results reveal a significant influence on the translation quality when using the wrong domain information, particularly in the News and e-Commerce domains. This aligns with previous observations where the correct domain information is used, indicating that domain information explicitly impacts ChatGPT on translations, especially on the domains with more specific words.

\subsection{Few-shot Prompts}

\begin{figure}[t]
\includegraphics[width=0.5\textwidth]{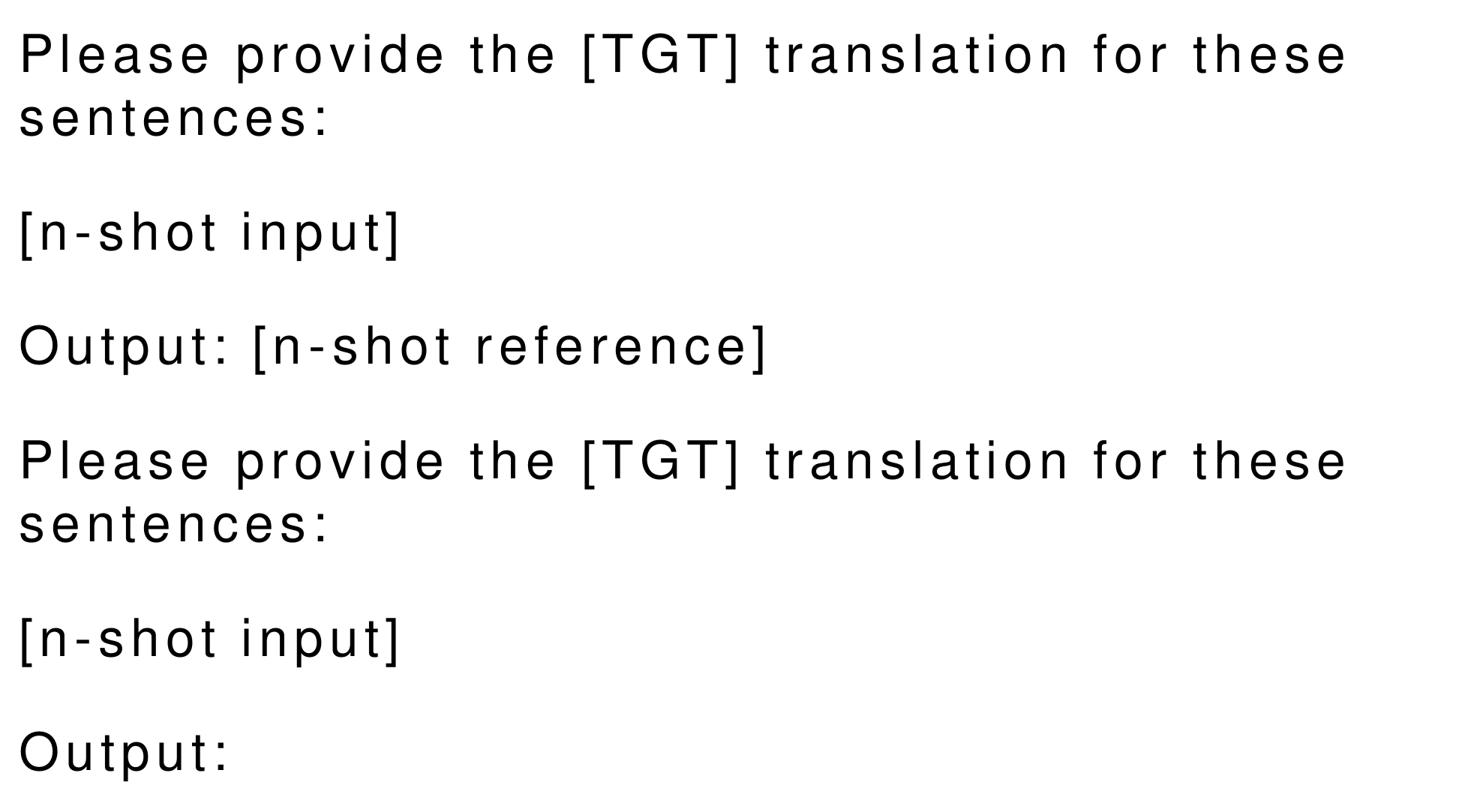}
\caption{The template of few-shot prompts. We take TP3 as an example.}
\label{fig4}
\end{figure}

Recent research \cite{brown2020language,chen2021revisiting,dou2022gpt} has demonstrated the advantages of in-context learning for LLMs. In-context learning involves adding several input-output examples (perform as prompts) to the input text to enhance the performance of LLMs across multiple tasks, without any adjustments to parameters or architecture. Li and Liang \shortcite{li2021prefix} proposed prefix-tuning that uses a few labelled data to enable general LLMs to achieve competitive results with a specialized vector trained on the downstream datasets. Inspired by prefix-tuning, Tsimpoukelli et al. \shortcite{tsimpoukelli2021multimodal} utilize in-context learning to improve LLM performance on a variety of multi-modal tasks. In addition, there is a series of research focused on prompting LLMs for MT \cite{vilar2022prompting,zhang2023prompting}. In-context learning relies on the quality and quantity of examples selected, with previous research suggesting that providing more high-quality examples leads to greater downstream task improvements \cite{yang2020survey,wei2022chain}. However, the number of examples is limited by LLM architecture \cite{olmo2021gpt3}, and providing more examples does not result in any meaningful improvement \cite{hendy2023good}.

\begin{figure}[t]
\includegraphics[width=0.5\textwidth]{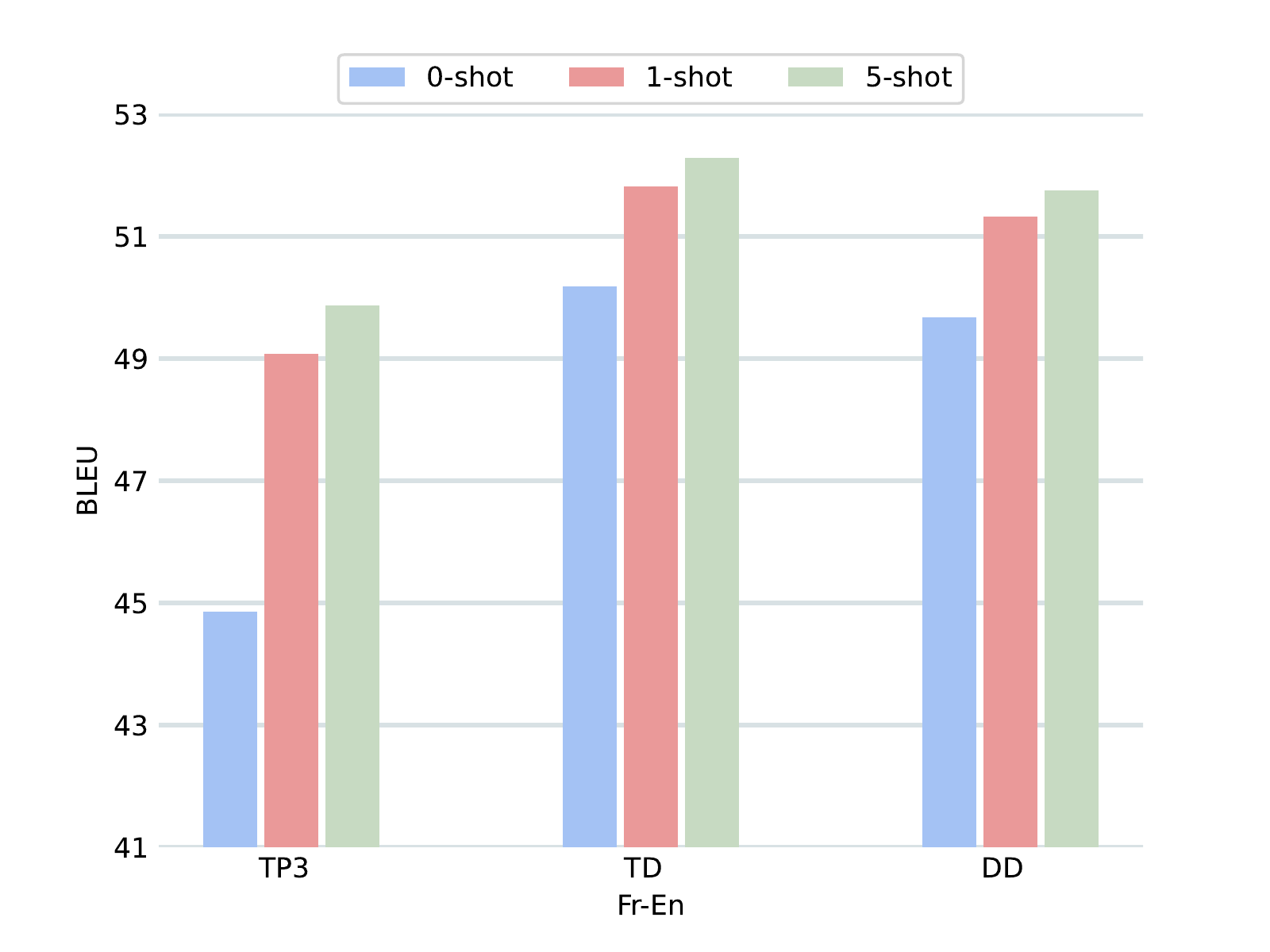}
\caption{Comparing BLEU scores of ChatGPT with 0, 1, 5-shot prompts on the Fr$\rightarrow$En translation.}
\label{fig2}
\end{figure}

\begin{figure}[t]
\includegraphics[width=0.5\textwidth]{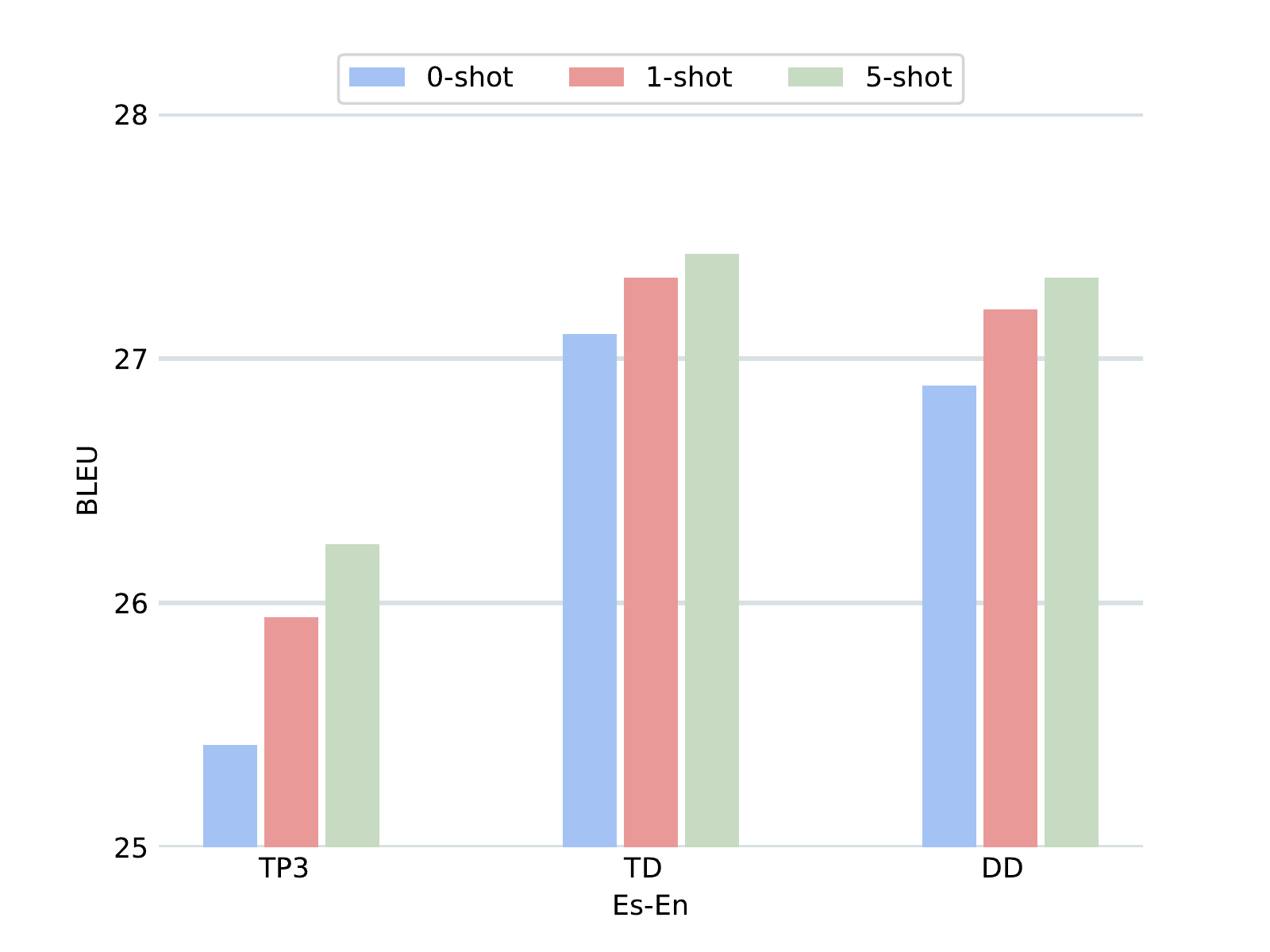}
\caption{Comparing BLEU scores of ChatGPT with 0, 1, 5-shot prompts on the Es$\rightarrow$En translation.}
\label{fig3}
\end{figure}

Thus, in this section, we perform experiments of few-shot prompts by selecting 1 and 5 examples (shots) from the original test sets and integrating them with the previously used prompts (i.e., TP3, TT, CD) as shown in Figure \ref{fig4}. Note that the selected shots do not overlap with the data used for testing. To access the quality of the shots, we use LaBSE \cite{feng2022language} to score the translation pairs following \cite{hendy2023good}, and then select the top-1 and top-5 pairs as in-context shots. Our experiments with few-shot prompts are conducted on French$\rightarrow$English and Spanish$\rightarrow$English, and the results are presented in Figure \ref{fig2}, \ref{fig3}. We compare the performance of three few-shot settings, namely 0-shot, 1-shot and 5-shot.

Based on the observations, ChatGPT with 1-shot and 5-shot setups result in substantial improvement in performance over 0-shot for TP3, TT and CD. However, it is worth noting that few-shot prompts yield the most significant gains when applied to the basic TP3, while the improvements are relatively limited for TT and CD. This observation is consistent in both French$\rightarrow$English and Spanish$\rightarrow$English translations. In other words, combining few-shot learning with designed prompts does not necessarily result in equivalent additive improvements. We speculate that there is some overlap between the information contained in in-context shots and the information provided by the designed prompts. On the other hand, the consistent improvement

\section{Conclusion and Further work}

In this work, we present an empirical study on how to unleash the translation power of ChatGPT for machine translations by using different translation prompts. Specifically, we evaluate and analyze our proposed prompts in various translation settings, including multilingual, multi-reference, multi-domain and few-shot translations. 

Our findings indicate that ChatGPT is capable of achieving better results than commercial translation systems by using properly designed prompts. In addition, we incorporate POS tags into TT and CD as auxiliary information, but the observed instabilities in partial translation directions indicate that this strategy requires further investigation to fully understand its limitations and potential.

Considering that ChatGPT is trained specifically for dialogue and prioritizes coherence and conciseness in generating sentences, which may not be sufficient to evaluate the translation quality with a single reference. We use the multi-reference test sets to evaluate the performance of ChatGPT. This allows us to take into account the diversity of possible translations. The experimental results indicate that the translation performance of ChatGPT is significantly different when evaluated with multiple references compared to the single reference evaluation. This highlights the importance of using more comprehensive evaluation criteria to assess the quality of ChatGPT's translations.

To further investigate the impact of CD on ChatGPT, we conducted experiments on multi-domain test sets. The experimental results show that introducing correct domain information into prompts can effectively improve the performance of ChatGPT for translation. Additionally, we use few-shot prompts to prompt ChatGPT, which can achieve substantial improvement over different translation directions.

In summary, we find that providing ChatGPT with the correct information about the input text (such as translation tasks and context domain) in the prompt can further unleash the performance of ChatGPT, however, when the information is too complex or noisy, it will produce severe performance degradation. In addition, using the few-shot example is a method that needs to be considered seriously, because input-output examples contain a lot of hidden information that cannot be explicitly conveyed through specific text.


\section*{Acknowledgements}

This work is supported by the 2020 Catalyst: Strategic New Zealand - Singapore Data Science Research Programme Fund by Ministry of Business, Innovation and Employment(MBIE), New Zealand.

\bibliographystyle{eamt23}
\bibliography{eamt23}

\begin{thebibliography}{}

\bibitem[\protect\citename{Brown \bgroup et al.\egroup
  }2020]{brown2020language}
Brown, Tom, Benjamin Mann, Nick Ryder, Melanie Subbiah, Jared~D Kaplan,
  Prafulla Dhariwal, Arvind Neelakantan, Pranav Shyam, Girish Sastry, Amanda
  Askell, et~al.
\newblock 2020.
\newblock Language models are few-shot learners.
\newblock {\em Advances in neural information processing systems}.

\bibitem[\protect\citename{Chen \bgroup et al.\egroup
  }2021]{chen2021revisiting}
Chen, Yiming, Yan Zhang, Chen Zhang, Grandee Lee, Ran Cheng, and Haizhou Li.
\newblock 2021.
\newblock Revisiting self-training for few-shot learning of language model.
\newblock {\em arXiv preprint arXiv:2110.01256}.

\bibitem[\protect\citename{Dou \bgroup et al.\egroup }2022]{dou2022gpt}
Dou, Yao, Maxwell Forbes, Rik Koncel-Kedziorski, Noah~A Smith, and Yejin Choi.
\newblock 2022.
\newblock Is gpt-3 text indistinguishable from human text? scarecrow: A
  framework for scrutinizing machine text.
\newblock In {\em ACL 2022}.

\bibitem[\protect\citename{Fan \bgroup et al.\egroup }2021]{fan2021beyond}
Fan, Angela, Shruti Bhosale, Holger Schwenk, Zhiyi Ma, Ahmed El-Kishky,
  Siddharth Goyal, Mandeep Baines, Onur Celebi, Guillaume Wenzek, Vishrav
  Chaudhary, et~al.
\newblock 2021.
\newblock Beyond english-centric multilingual machine translation.
\newblock {\em The Journal of Machine Learning Research}.

\bibitem[\protect\citename{Feng \bgroup et al.\egroup }2022]{feng2022language}
Feng, Fangxiaoyu, Yinfei Yang, Daniel Cer, Naveen Arivazhagan, and Wei Wang.
\newblock 2022.
\newblock Language-agnostic bert sentence embedding.
\newblock In {\em ACL 2022}.

\bibitem[\protect\citename{Goyal \bgroup et al.\egroup }2022]{goyal2022flores}
Goyal, Naman, Cynthia Gao, Vishrav Chaudhary, Peng-Jen Chen, Guillaume Wenzek,
  Da~Ju, Sanjana Krishnan, Marc’Aurelio Ranzato, Francisco Guzm{\'a}n, and
  Angela Fan.
\newblock 2022.
\newblock The flores-101 evaluation benchmark for low-resource and multilingual
  machine translation.
\newblock {\em Transactions of the Association for Computational Linguistics}.

\bibitem[\protect\citename{Guo \bgroup et al.\egroup
  }2020]{guo2020incorporating}
Guo, Junliang, Zhirui Zhang, Linli Xu, Hao-Ran Wei, Boxing Chen, and Enhong
  Chen.
\newblock 2020.
\newblock Incorporating bert into parallel sequence decoding with adapters.
\newblock {\em Advances in Neural Information Processing Systems}.

\bibitem[\protect\citename{Hendy \bgroup et al.\egroup }2023]{hendy2023good}
Hendy, Amr, Mohamed Abdelrehim, Amr Sharaf, Vikas Raunak, Mohamed Gabr,
  Hitokazu Matsushita, Young~Jin Kim, Mohamed Afify, and Hany~Hassan Awadalla.
\newblock 2023.
\newblock How good are gpt models at machine translation? a comprehensive
  evaluation.
\newblock {\em arXiv preprint arXiv:2302.09210}.

\bibitem[\protect\citename{Jiao \bgroup et al.\egroup }2023]{jiao2023chatgpt}
Jiao, Wenxiang, Wenxuan Wang, Jen-tse Huang, Xing Wang, and Zhaopeng Tu.
\newblock 2023.
\newblock Is chatgpt a good translator? a preliminary study.
\newblock {\em arXiv preprint arXiv:2301.08745}.

\bibitem[\protect\citename{Khan \bgroup et al.\egroup }2020]{khan2020novel}
Khan, Nabeel~Sabir, Adnan Abid, and Kamran Abid.
\newblock 2020.
\newblock A novel natural language processing (nlp)--based machine translation
  model for english to pakistan sign language translation.
\newblock {\em Cognitive Computation}.

\bibitem[\protect\citename{Lewis \bgroup et al.\egroup }2020]{lewis2020bart}
Lewis, Mike, Yinhan Liu, Naman Goyal, Marjan Ghazvininejad, Abdelrahman
  Mohamed, Omer Levy, Veselin Stoyanov, and Luke Zettlemoyer.
\newblock 2020.
\newblock Bart: Denoising sequence-to-sequence pre-training for natural
  language generation, translation, and comprehension.
\newblock In {\em ACL 2020}.

\bibitem[\protect\citename{Li and Liang}2021]{li2021prefix}
Li, Xiang~Lisa and Percy Liang.
\newblock 2021.
\newblock Prefix-tuning: Optimizing continuous prompts for generation.
\newblock In {\em ACL 2021}.

\bibitem[\protect\citename{Li \bgroup et al.\egroup }2022]{li2022prompt}
Li, Yafu, Yongjing Yin, Jing Li, and Yue Zhang.
\newblock 2022.
\newblock Prompt-driven neural machine translation.
\newblock In {\em Findings of the Association for Computational Linguistics:
  ACL 2022}.

\bibitem[\protect\citename{Liu \bgroup et al.\egroup
  }2019]{liu2019incorporating}
Liu, Zihan, Yan Xu, Genta~Indra Winata, and Pascale Fung.
\newblock 2019.
\newblock Incorporating word and subword units in unsupervised machine
  translation using language model rescoring.
\newblock In {\em WMT 2019}.

\bibitem[\protect\citename{Niu and Carpuat}2020]{niu2020controlling}
Niu, Xing and Marine Carpuat.
\newblock 2020.
\newblock Controlling neural machine translation formality with synthetic
  supervision.
\newblock In {\em AAAI 2020}.

\bibitem[\protect\citename{Olmo \bgroup et al.\egroup }2021]{olmo2021gpt3}
Olmo, Alberto, Sarath Sreedharan, and Subbarao Kambhampati.
\newblock 2021.
\newblock Gpt3-to-plan: Extracting plans from text using gpt-3.
\newblock {\em arXiv preprint arXiv:2106.07131}.

\bibitem[\protect\citename{Ott \bgroup et al.\egroup }2018]{ott2018analyzing}
Ott, Myle, Michael Auli, David Grangier, and Marc’Aurelio Ranzato.
\newblock 2018.
\newblock Analyzing uncertainty in neural machine translation.
\newblock In {\em PLMR}.

\bibitem[\protect\citename{Papineni \bgroup et al.\egroup
  }2002]{papineni2002bleu}
Papineni, Kishore, Salim Roukos, Todd Ward, and Wei-Jing Zhu.
\newblock 2002.
\newblock Bleu: a method for automatic evaluation of machine translation.
\newblock In {\em ACL 2002}.

\bibitem[\protect\citename{Peng \bgroup et al.\egroup }2023]{peng2023towards}
Peng, Keqin, Liang Ding, Qihuang Zhong, Li~Shen, Xuebo Liu, Min Zhang, Yuanxin
  Ouyang, and Dacheng Tao.
\newblock 2023.
\newblock Towards making the most of chatgpt for machine translation.
\newblock {\em arXiv preprint arXiv:2303.13780}.

\bibitem[\protect\citename{Perera \bgroup et al.\egroup
  }2022]{perera2022improving}
Perera, Ravinga, Thilakshi Fonseka, Rashmini Naranpanawa, and Uthayasanker
  Thayasivam.
\newblock 2022.
\newblock Improving english to sinhala neural machine translation using
  part-of-speech tag.
\newblock {\em arXiv preprint arXiv:2202.08882}.

\bibitem[\protect\citename{Petrushkov \bgroup et al.\egroup
  }2018]{petrushkov2018learning}
Petrushkov, Pavel, Shahram Khadivi, and Evgeny Matusov.
\newblock 2018.
\newblock Learning from chunk-based feedback in neural machine translation.
\newblock In {\em ACL 2018 (Volume 2: Short Papers)}.

\bibitem[\protect\citename{Popovi{\'c}}2017]{popovic2017chrf++}
Popovi{\'c}, Maja.
\newblock 2017.
\newblock chrf++: words helping character n-grams.
\newblock In {\em Proceedings of the second conference on machine translation}.

\bibitem[\protect\citename{Post}2018]{post2018call}
Post, Matt.
\newblock 2018.
\newblock A call for clarity in reporting bleu scores.
\newblock In {\em Proceedings of the Third Conference on Machine Translation:
  Research Papers}.

\bibitem[\protect\citename{Qi \bgroup et al.\egroup }2020]{qi2020stanza}
Qi, Peng, Yuhao Zhang, Yuhui Zhang, Jason Bolton, and Christopher~D Manning.
\newblock 2020.
\newblock Stanza: A python natural language processing toolkit for many human
  languages.
\newblock In {\em ACL 2020: System Demonstrations}.

\bibitem[\protect\citename{Snover \bgroup et al.\egroup }2006]{snover2006study}
Snover, Matthew, Bonnie Dorr, Richard Schwartz, Linnea Micciulla, and John
  Makhoul.
\newblock 2006.
\newblock A study of translation edit rate with targeted human annotation.
\newblock In {\em AMTA 2006}.

\bibitem[\protect\citename{Stahlberg}2020]{stahlberg2020neural}
Stahlberg, Felix.
\newblock 2020.
\newblock Neural machine translation: A review.
\newblock {\em Journal of Artificial Intelligence Research}.

\bibitem[\protect\citename{Tsimpoukelli \bgroup et al.\egroup
  }2021]{tsimpoukelli2021multimodal}
Tsimpoukelli, Maria, Jacob~L Menick, Serkan Cabi, SM~Eslami, Oriol Vinyals, and
  Felix Hill.
\newblock 2021.
\newblock Multimodal few-shot learning with frozen language models.
\newblock {\em Advances in Neural Information Processing Systems}.

\bibitem[\protect\citename{Vilar \bgroup et al.\egroup
  }2022]{vilar2022prompting}
Vilar, David, Markus Freitag, Colin Cherry, Jiaming Luo, Viresh Ratnakar, and
  George Foster.
\newblock 2022.
\newblock Prompting palm for translation: Assessing strategies and performance.
\newblock {\em arXiv preprint arXiv:2211.09102}.

\bibitem[\protect\citename{Wei \bgroup et al.\egroup }2022]{wei2022chain}
Wei, Jason, Xuezhi Wang, Dale Schuurmans, Maarten Bosma, Ed~Chi, Quoc Le, and
  Denny Zhou.
\newblock 2022.
\newblock Chain of thought prompting elicits reasoning in large language
  models.
\newblock {\em arXiv preprint arXiv:2201.11903}.

\bibitem[\protect\citename{Yang \bgroup et al.\egroup }2019]{yang2019xlnet}
Yang, Zhilin, Zihang Dai, Yiming Yang, Jaime Carbonell, Russ~R Salakhutdinov,
  and Quoc~V Le.
\newblock 2019.
\newblock Xlnet: Generalized autoregressive pretraining for language
  understanding.
\newblock {\em Advances in neural information processing systems}.

\bibitem[\protect\citename{Yang \bgroup et al.\egroup }2020]{yang2020survey}
Yang, Shuoheng, Yuxin Wang, and Xiaowen Chu.
\newblock 2020.
\newblock A survey of deep learning techniques for neural machine translation.
\newblock {\em arXiv preprint arXiv:2002.07526}.

\bibitem[\protect\citename{Zhang \bgroup et al.\egroup
  }2023]{zhang2023prompting}
Zhang, Biao, Barry Haddow, and Alexandra Birch.
\newblock 2023.
\newblock Prompting large language model for machine translation: A case study.
\newblock {\em arXiv preprint arXiv:2301.07069}.

\end{thebibliography}

\end{document}